\documentclass[journal]{IEEEtran}

\usepackage{amsmath} 
\usepackage{amssymb}
\usepackage{amsmath}

\usepackage{amssymb}
\usepackage{amsfonts}
\usepackage{float}
\UseRawInputEncoding
\usepackage{textcomp}
\usepackage{graphicx}
\usepackage[utf8]{inputenc}
\usepackage{array}
\usepackage{booktabs}
\usepackage{multirow}
\usepackage{colortbl}
\usepackage{threeparttable}
\usepackage{xcolor} 
\usepackage{CJKutf8}

\usepackage[linesnumbered,ruled,vlined]{algorithm2e}

\usepackage{subcaption} 

\usepackage{lipsum} 
\usepackage{url}

\usepackage[noadjust]{cite} 

\usepackage[colorlinks=true, linkcolor=blue, citecolor=green, urlcolor=red]{hyperref}
\usepackage{multirow}
\usepackage{booktabs}
\DeclareUnicodeCharacter{21BB}{\ensuremath{\circlearrowright}}
\begin{document}

\author{
\IEEEauthorblockN{Mingao Tan$^{1,2\dagger}$, Yiyang Li$^{1,2\dagger}$, Shanze Wang$^{2}$, Xinming Zhang$^{2}$, Wei Zhang$^{2*}$\\}
\IEEEauthorblockA{${}^{1}$ School of Computer Science, Shanghai Jiao Tong University, Shanghai, P. R. China\\
${}^{2}$ College of Information Science and Technology, Eastern Institute of Technology, Ningbo, P. R. China\\
${}^{\dagger}$ These authors contributed equally to this work.\\
${}^{*}$ Corresponding author.}
}

\title{\LARGE \bf FSUNav: A Cerebrum-Cerebellum Architecture for Fast, Safe, and Universal Zero-Shot Goal-Oriented Navigation}


\maketitle

\begin{abstract}

Current vision-language navigation methods face substantial bottlenecks regarding heterogeneous robot compatibility, real-time performance, and navigation safety. Furthermore, they struggle to support open-vocabulary semantic generalization and multimodal task inputs. To address these challenges, this paper proposes FSUNav: a Cerebrum-Cerebellum architecture for fast, safe, and universal zero-shot goal-oriented navigation, which innovatively integrates vision-language models (VLMs) with the proposed architecture. The cerebellum module, a high-frequency end-to-end module, develops a universal local planner based on deep reinforcement learning, enabling unified navigation across heterogeneous platforms (e.g., humanoid, quadruped, wheeled robots) to improve navigation efficiency while significantly reducing collision risk. The cerebrum module constructs a three-layer reasoning model and leverages VLMs to build an end-to-end detection and verification mechanism, enabling zero-shot open-vocabulary goal navigation without predefined IDs and improving task success rates in both simulation and real-world environments. Additionally, the framework supports multimodal inputs (e.g., text, target descriptions, and images), further enhancing generalization, real-time performance, safety, and robustness. Experimental results on MP3D, HM3D, and OVON benchmarks demonstrate that FSUNav achieves state-of-the-art performance on object, instance image, and task navigation, significantly outperforming existing methods. Real-world deployments on diverse robotic platforms further validate its robustness and practical applicability.

\end{abstract}


%
\IEEEpeerreviewmaketitle

\section{Introduction}
Vision-Language Navigation (VLN) is a core task in Embodied Artificial Intelligence, aiming to enable robots to autonomously navigate to targets based on natural language instructions in visual environments. \cite{Anderson2017VisionandLanguageNI} While deep learning and pre-trained models have driven rapid progress in this field, significant bottlenecks remain when transferring these advances to real-world robotic systems.\cite{Gu2022VisionandLanguageNA} The primary challenge lies in the compatibility of heterogeneous robots: most existing methods are designed for specific platforms (e.g., wheeled robots) and struggle to generalize to humanoid, quadruped, or other heterogeneous platforms, requiring extensive retraining and adaptation for cross-platform use.\cite{580977} \cite{6309484} Second, insufficient real-time performance limits the practicality of VLN in dynamic environments. \cite{Anderson2017VisionandLanguageNI}\cite{Moudgil2021SOATAS} \cite{Hong2020VLNBERTAR} Many state-of-the-art approaches rely on computationally intensive large models with high inference latency, making it difficult to meet the low-latency requirements of robotic navigation. This issue is particularly critical in fast-changing scenarios such as crowded indoor spaces or cluttered outdoor environments, where delayed decision-making can lead to navigation failures or even safety incidents. Meanwhile, existing research has largely focused on path planning and success rate optimization in simulation, with insufficient attention to safety mechanisms such as collision avoidance and dynamic obstacle avoidance. \cite{Krantz2020BeyondTN} \cite{Gu2022VisionandLanguageNA} In the real world, robots must contend with pedestrians, moving objects, narrow corridors, and other uncertainties, making efficient movement with active collision avoidance essential for practical deployment. Furthermore, current VLN systems exhibit clear limitations in semantic generalization and multimodal interaction. Most methods rely on predefined object categories or fixed vocabularies for target recognition, failing to understand open-vocabulary instructions and thus limiting their applicability in open-world settings. Moreover, task inputs are often restricted to a single modality (e.g., text), overlooking the auxiliary role of visual (e.g., target images) and auditory information in navigation, which hinders the ability to meet the rich demands of human-robot interaction in complex tasks \cite{Majumdar2022ZSONZO}\cite{NEURIPS2022_27c546ab}.

To address the challenges mentioned above, this paper proposes FSUNav: a Cerebrum-Cerebellum Architecture for Fast, Safe, and Universal Zero-Shot Goal-Oriented Navigation. We evaluate the proposed framework on three tasks—Object Goal \cite{chaplot2020object}\cite{Zhou2023ESCEW}\cite{yin2024sg}, Instance-Image Goal \cite{krantz2022instance}, and Text Goal navigation—using the MP3D \cite{Matterport3D} and HM3D \cite{Ramakrishnan2021HabitatMatterport3D} datasets, as well as real-world experiments within \cite{Sun2024PrioritizedSL}. The results demonstrate that the brain module of our method achieves consistent performance improvements on fixed datasets and effectively unifies multiple navigation tasks. In complex real-world environments, the large and small brain modules enable unified navigation for heterogeneous robots, including humanoid, quadruped, and wheeled robots, significantly enhancing navigation efficiency while reducing collision risk. It provides a systematic solution to the core bottlenecks. 

The main contributions are reflected in the following aspects:

\begin{itemize}
\item Cerebellum Module (Low-Level Execution and Control): A robot-platform-decoupled local planning module based on deep reinforcement learning, featuring dimension-configurable input representations and a novel learning curriculum. It enables generalization across heterogeneous robots (humanoid, quadruped, wheeled) with varying physical sizes, speed ranges, and acceleration constraints, achieving fast and safe local navigation control.
\item Cerebrum Module (High-Level Understanding and Decision-Making): A VLM-based module for high-level semantic understanding and goal verification, supporting natural language or visual target inputs. Through cross-modal alignment, it achieves open-vocabulary zero-shot target localization without predefined categories. Simulation and real-world experiments show that this mechanism significantly improves task success in open-vocabulary scenarios. A unified architecture comprising semantic, spatial, and rule layers is constructed to support three tasks—Object-Goal Navigation (ON), Instance-Image Navigation (IIN), and Text-Goal Navigation (TN)—under a unified reasoning pipeline without task-specific training.
\item By combining the Cerebrum and Cerebellum modules, the framework achieves three major goals: cross-platform compatibility, real-time safe navigation, and open-vocabulary semantic generalization. Systematic experiments in diverse simulation environments and on real robots validate its generality across heterogeneous platforms, safety and efficiency in complex scenes, and successful zero-shot multi-modal task execution. These results provide theoretical and technical foundations for the practical deployment of vision-language navigation in complex dynamic environments.
\end{itemize}






\section{Related work}

\subsection{Zero-shot Vision-Language Navigation}
Current goal-oriented navigation methods are mainly divided into two categories: supervised learning and zero-shot learning. Supervised learning methods typically follow two technical approaches: one is end-to-end training of visual encoders using reinforcement learning or agent policies; the other is constructing structured semantic maps from annotated data \cite{Anderson2017VisionandLanguageNI}\cite{Chaplot2020NeuralTS}. These methods perform well in known environments, but their performance heavily depends on the distribution of the training data. When faced with unseen object categories or spatial layouts, their generalization capability significantly degrades. To overcome this limitation, recent zero-shot navigation methods no longer rely on task-specific training data.\cite{Yin2025UniGoalTU} Among these, image-based approaches achieve target localization by mapping detected objects into a visual-semantic embedding space. Map-based methods mainly consist of two paradigms: one builds exploration maps based on frontiers, \cite{Yokoyama2023VLFMVF}while the other constructs waypoint-based map systems and leverages vision-language models for joint reasoning, enabling more flexible navigation decisions in unknown environments.\cite{Nie2025WMNavIV}

\subsection{Cross-Embodiment and Dimension-Configurable Planning}

Conventional navigation planners, such as DWA\cite{580977} and TEB\cite{6309484}, suffer from tight coupling with specific robot kinematics, which restricts their generalization across different platforms. To mitigate this, recent research has focused on embodiment-agnostic frameworks. In the context of LiDAR-based navigation, DRL-DCLP\cite{10900448} has demonstrated success in zero-shot adaptation for rectangular differential-drive robots of varying sizes. However, Vision-Language Navigation currently lacks a rapid and secure visual navigation approach capable of accommodating variations in robot morphology. To bridge this gap, we propose FSUNav: a Cerebrum-Cerebellum Architecture for Fast, Safe, and Universal Zero-Shot Goal-Oriented Navigation. Our method manages diverse robot dimensions in a zero-shot manner, thereby eliminating the need for platform-specific data collection or post-deployment fine-tuning.

\section{Problem Definition}
The navigation system in this paper adopts a hierarchical architecture consisting of a Cerebrum module and a Cerebellum module. The Cerebrum module performs high-level reasoning and generates a two-dimensional target coordinate $\mathbf{g} \in \mathbb{R}^2$ on a pre-built environmental map. The Cerebellum module, relying solely on its own odometry information, executes real-time mapless navigation and obstacle avoidance to drive the robot toward the specified target. Success is defined as the robot calling the \texttt{stop} action when it reaches within a Euclidean distance $d$ of the target point $\mathbf{g}$.

Fast and safe universal zero-shot goal-oriented navigation introduces four key challenges beyond the standard goal-oriented navigation task:

\begin{enumerate}
    \item \textbf{Multimodal Unification}: The agent must be capable of interpreting and executing navigation commands expressed in any predefined goal modality—including object categories, instance images, and natural language instructions—using a single unified policy \cite{Chang2023GOATGT}.
    \item \textbf{Cross-Platform Generalization}: The policy must generalize across diverse heterogeneous robotic platforms, adaptively handling variations in kinematic models, velocity limits, and acceleration configurations, thereby enabling truly platform-agnostic deployment.
    \item \textbf{Novel Scene Adaptation}: The agent must perform effectively in previously unseen environments and on novel goal specifications without any task-specific training or fine-tuning, relying solely on its pre-trained or inherently generalizable capabilities \cite{Yin2025UniGoalTU}.
    \item \textbf{Efficiency and Safety}: The agent should achieve efficient task completion during navigation while maintaining precise obstacle avoidance capabilities, thereby improving navigation efficiency and effectively reducing collision risk to ensure safe and reliable execution.
\end{enumerate}

\section{Method}

\subsection{Overall framework of FSUNav}

FSUNav is a fast, safe, universal zero-shot goal-oriented navigation method based on a cerebrum-cerebellum architecture. The Cerebellum module is a deep reinforcement learning local planner decoupled from robotic platforms, enabling fast, safe control across heterogeneous robots. The Cerebrum module uses a three-layer structure—Semantic, Spatial, and Rule—unifying semantic understanding, spatial reasoning, and decision-making. It leverages a Vision-Language Model (VLM) to support natural language or visual goals, enabling open-vocabulary zero-shot localization and unifying object-goal, instance-image, and text-goal navigation. Operating in closed loop, the cerebrum generates high-level intentions while the cerebellum executes precise motion with collision avoidance, achieving cross-platform compatibility and open-vocabulary generalization without sacrificing real-time performance or safety.

\begin{figure*}
    \centering
    \includegraphics[width=0.8\linewidth]{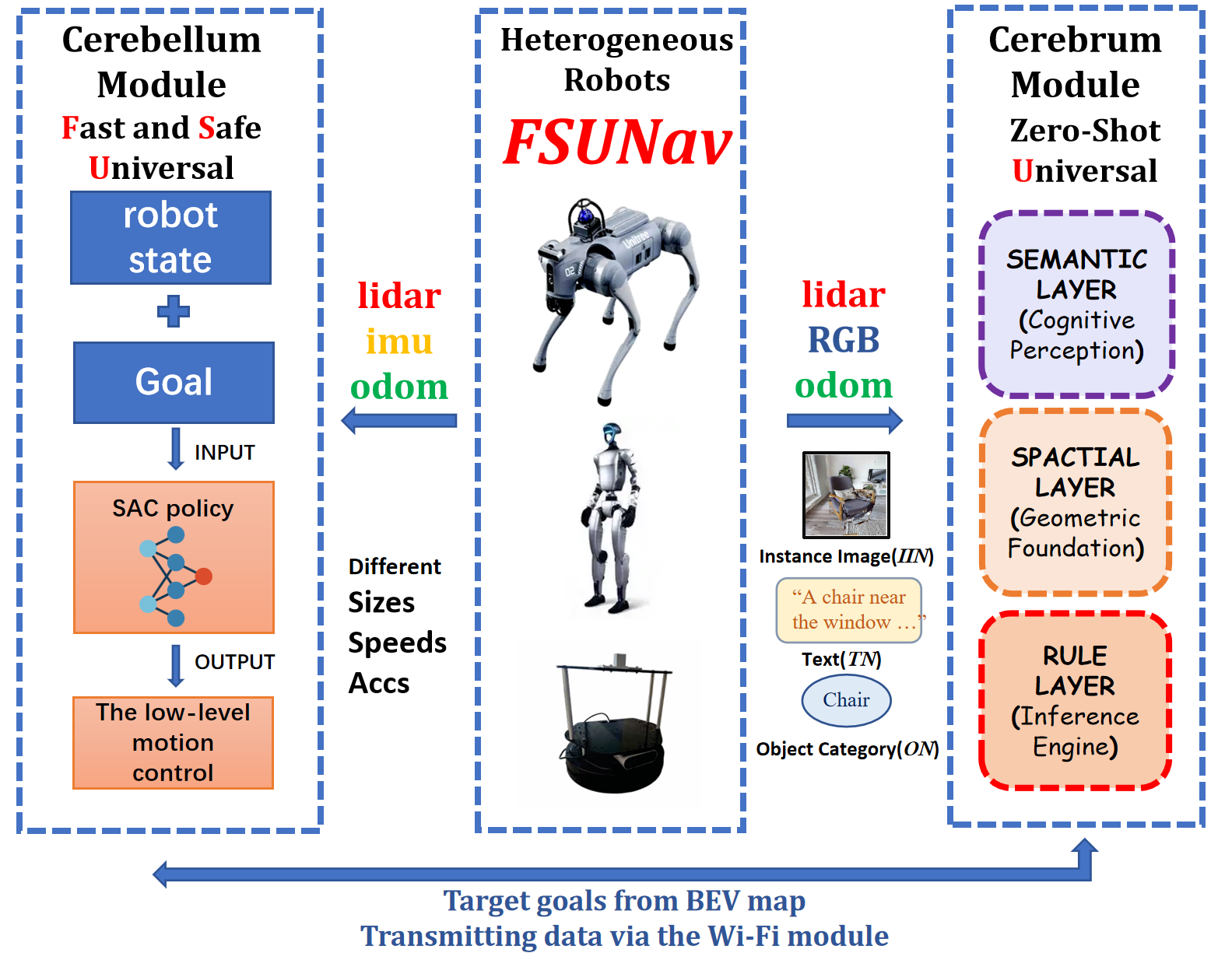}
    \caption{The overall framework of FSUNav is shown in the figure. Current vision-language navigation methods still face significant limitations in terms of heterogeneous robot compatibility, real-time performance, and navigation safety, while also struggling to support open-vocabulary semantic generalization and multimodal task inputs. To address these issues, this paper proposes FSUNav, which leverages an efficient dual-brain collaborative architecture to achieve fast, safe, and generalizable zero-shot goal-oriented navigation, delivering comprehensive improvements in generalization capability, real-time performance, safety, and robustness. }
    \label{fig:placeholder}
\end{figure*}

\subsection{Cerebellum Module}

The Cerebellum module frames local navigation as a deep reinforcement learning (DRL) decision-making problem, removing the need for large datasets, platform-specific fine-tuning, or retraining. It yields real-time velocity commands \( \mathbf{u}_\mathrm{t} = (v_\mathrm{t}, \omega_\mathrm{t}) \) for target‑oriented obstacle avoidance using only local observations. 

To achieve generalization across diverse robotic platforms, the approach integrates a dimension‑configurable policy network with a curriculum learning framework. At each discrete time step \( t \), the policy input \( \mathbf{o}_\mathrm{t} \) aggregates a unified geometric representation, the robot state, and a relative goal:
\begin{equation}
\mathbf{o}_\mathrm{t} = \{ \mathbf{xr}_\mathrm{t},\ \mathbf{g}_\mathrm{t},\ \mathbf{v}_\mathrm{t},\ \mathbf{L}_{\text{dyn}} \},
\end{equation}
where 
\( \mathbf{xr}_\mathrm{t} \) is the unified geometric representation,
\( \mathbf{g}_\mathrm{t} \in \mathbb{R}^2 \) is the relative goal position expressed in the robot’s local frame,
\( \mathbf{v}_\mathrm{t} = (v_\mathrm{t}, \omega_\mathrm{t}) \) denotes the current velocity,
and \( \mathbf{L}_{\text{dyn}} \) contains the velocity limits (\( v_{\max}, \omega_{\max} \)) and acceleration limits (\( a_{\mathbf{v},\max}, a_{\omega,\max} \)).

Crucially, the unified geometric representation \( \mathbf{xr}_\mathrm{t} \) is processed as a set of points \( \mathbf{xr}_\mathrm{t} = \{ \mathbf{p}_{\mathrm{t,i}} \}_{i=1}^n \). This representation consolidates heterogeneous sensor inputs—such as depth cameras or LiDAR—into a consistent geometric format, enabling seamless deployment across robots with different sensing modalities.

To explicitly incorporate the robot’s geometry into the decision‑making process, the robot is modeled as a cuboid with dimensions \( \mathbf{body} = [L_{\text{front}},\ L_{\text{rear}},\ W] \), where \( L_{\text{front}} \) and \( L_{\text{rear}} \) denote the distances from the drive center to the front and rear extremities, respectively, and \( W \) is the total width. Each obstacle point is then encoded with dimension‑aware features:
\begin{equation}
\mathbf{p}_{\mathrm{t,i}} = \left( \sin \phi_{\mathrm{t,i}},\ \cos \phi_{\mathrm{t,i}},\ \frac{1}{d_{\mathrm{t,i}} - \beta},\ L_{\text{front}},\ L_{\text{rear}},\ \frac{W}{2} \right),
\end{equation}
where \( d_{\mathrm{t,i}} \) and \( \phi_{\mathrm{t,i}} \) are the distance and angle of the \( i \)-th scan point, and \( \beta \) is a trainable scaling parameter. This formulation directly embeds the robot’s physical dimensions into each obstacle point, enabling the policy network to assess obstacle criticality based on its own geometric constraints.

The point set is processed by a PointNet encoder: each point is independently transformed by a shared multi‑layer perceptron (MLP), followed by a max‑pooling operation to aggregate a compact global geometric feature. This feature is then concatenated with the goal and state information, and the final action is generated by a Soft Actor‑Critic (SAC) network.

We employ a curriculum learning strategy in simulation to train the policy over a continuous space of robot configurations. Specifically, robots are grouped by their perimeter, with each group assigned to a progressively more challenging training scenario. Initial scenarios feature open maps, while subsequent scenarios gradually shrink the navigable area to increase difficulty. At each stage, if the success rate of a group in the current scenario exceeds a predefined threshold (e.g., 90\%), all robots in that group are promoted to the next scenario. This staged progression allows the policy to incrementally learn adaptive obstacle avoidance behavior across varying robot dimensions. As a result, the Cerebellum module achieves \textbf{zero-shot cross-embodiment deployment}. Because the input representation explicitly encodes robot dimensions and dynamic limits, and PointNet naturally handles variable-sized point sets, the trained policy generalizes seamlessly to both wheeled mobile robots and quadruped robots of varying sizes and sensor configurations (including LiDAR placement, field of view, and sampling density) without requiring any platform-specific retraining or fine-tuning.


\subsection{Cerebrum Module}

To enable a unified and training-free reasoning framework for heterogeneous goal-oriented navigation tasks, we propose a Three-Layer Cerebrum Module that explicitly decouples semantic understanding, spatial reasoning, and decision-making. This modular design allows each layer to operate with plug-and-play knowledge components, ensuring both interpretability and adaptability without relying on implicit end-to-end learning. In contrast to traditional pipelines that treat perception, planning, and control as separate modules, our architecture is built around a single Vision-Language Model (VLM) that serves as a unified semantic reasoning engine across all layers.

\paragraph{Semantic Layer.}
The Semantic Layer acts as the perceptual interface, leveraging the VLM to transform multimodal goal specifications---such as object categories, reference images, or natural language instructions---into a structured target profile. We define a parsing function $\Phi$ that prompts the VLM to output a unified representation:
\begin{equation}
\mathcal{P}_{\mathrm{target}} = \Phi(\mathbf{G}) = \langle c_{\mathrm{core}}, \mathcal{A}_{\mathrm{attr}}, \mathcal{R}_{\mathrm{ctx}} \rangle,
\end{equation}
where $c_{\mathrm{core}}$ denotes the core object category, $\mathcal{A}_{\mathrm{attr}}$ represents fine-grained visual attributes (e.g., color, material), and $\mathcal{R}_{\mathrm{ctx}}$ captures spatial or semantic relationships with surrounding objects. This profile is simultaneously used for open-vocabulary visual grounding: the VLM produces normalized bounding boxes in a structured JSON format, which are then mapped to pixel coordinates using the image dimensions. The resulting detections are projected into 3D space via depth data to form semantic nodes for subsequent verification. This approach eliminates the need for task-specific detectors and supports arbitrary goal descriptions at test time.

\paragraph{Spatial Layer.}
The Spatial Layer maintains a geometric representation of the environment to support efficient exploration. We construct a bird's-eye-view (BEV) map that classifies each cell into free, occupied, or unknown states based on depth observations. However, purely geometry-based frontier exploration often lacks semantic guidance. To address this, we introduce a VLM-driven semantic exploration mechanism: when the target is not visible, the VLM is prompted to select a point on the floor that is safe, reachable, and semantically likely to lead toward the goal. This point serves as an intermediate subgoal. Formally, given observation $\mathcal{I}_t$, the VLM outputs a pixel coordinate $p_t$ (or $\varnothing$ if none is suitable), which is back-projected to a 3D waypoint $\mathbf{w}_t$ using depth data. The agent navigates to $\mathbf{w}_t$, injecting semantic priors into the exploration policy and reducing wasteful random steps. Concurrently, we extract frontier regions---boundaries between free and unknown space---and evaluate their utility using a geodesic distance field computed via the Fast Marching Method. The utility function for a frontier $f$ is defined as:
\begin{equation}
J(f) = \frac{\mathcal{I}(f)}{D(p_\mathrm{t}, f) + \epsilon} \cdot \mathbb{M}_{\mathrm{vis}}(f),
\end{equation}
where $\mathcal{I}(f)$ is the expected information gain, $D(p_\mathrm{t}, f)$ is the geodesic distance from the current pose $p_\mathrm{t}$, and $\mathbb{M}_{\mathrm{vis}}(f)$ is a binary mask that suppresses previously visited frontiers, thereby preventing redundant backtracking. The combination of VLM-generated semantic waypoints and geometry-based frontiers yields a hybrid exploration strategy that balances efficiency with semantic directedness.

\paragraph{Rule Layer.}
The Rule Layer serves as the central inference engine, coordinating the behavioral policy by alternating between exploration and confirmation modes. When the target is not in view, the agent executes the hybrid exploration strategy described above. Upon detecting a potential target, we employ a two-stage VLM-driven verification process. The first stage performs global scene verification: the VLM determines whether the target is present in the current view. If confirmed, the second stage conducts local instance identification by comparing candidate cropped regions against the target profile. Let $c$ denote a cropped image patch, and let $\mathcal{G}$ encode the goal specification; the verification function is defined as:
\begin{equation}
\mathcal{V}(c,\mathcal{G}) = \begin{cases}
1 & \text{if the VLM confirms a match},\\
0 & \text{otherwise}.
\end{cases}
\end{equation}
Only when $\mathcal{V}=1$ does the system commit the candidate to the global goal map and transition from exploration to goal-approaching behavior. To mitigate the computational overhead of frequent VLM queries, we enforce an adaptive cooldown mechanism:
\begin{equation}
\tau_{t+1} = 
\begin{cases}
T_\mathrm{l}, & \text{if target absent},\\
T_\mathrm{s}, & \text{if instance mismatch},\\
0, & \text{otherwise},
\end{cases}
\end{equation}
where $T_\mathrm{l}$ and $T_\mathrm{s}$ denote long and short cooldown periods, respectively. Additionally, the VLM contributes to building a semantic scene graph by inferring spatial relationships between objects (e.g., ``on'', ``under'', ``next to''), enriching the environment representation for higher-level reasoning. This hierarchical reasoning pipeline ensures robust target localization while maintaining computational efficiency.

Together, the three layers form a cohesive Cerebrum architecture centered around the VLM, enabling unified zero-shot navigation across object-, image-, and text-based goals without task-specific training.

\subsection{VLM is all your need}

A central innovation of our system lies in the \textbf{multi-level utilization of a Vision-Language Model (VLM)} beyond conventional object detection. Unlike prior work that treats VLMs as mere perception modules, we integrate the VLM as a \emph{semantic reasoning engine} that actively participates in goal interpretation, exploration guidance, verification, and scene understanding. This design enables the agent to handle open-vocabulary goals, reduce false positives, and navigate efficiently in unseen environments.

Instead of relying on a fixed set of object categories, we leverage the VLM to perform \emph{language-conditioned detection}. The goal can be specified in three forms: (i) an object category, (ii) a free-form textual description, or (iii) a reference image. For each, we construct a tailored prompt that elicits bounding boxes in a structured JSON format. The VLM outputs normalized coordinates, which are then mapped to pixel space:

\begin{equation}
\text{bbox}_{\text{px}} = \bigl[x_1\cdot W,\; y_1\cdot H,\; x_2\cdot W,\; y_2\cdot H\bigr],
\end{equation}

where \(W,H\) are the image dimensions. This approach eliminates the need for training a domain-specific detector and supports open-vocabulary goals at test time.

Beyond detection, the VLM contributes to exploration efficiency through a \emph{semantic waypoint generation} mechanism (detailed in the Spatial Layer), which provides semantically meaningful subgoals when the target is not visible. For robust target localization, a \emph{two-stage verification} process (described in the Rule Layer) filters false positives by combining global scene verification with local instance matching. Furthermore, the VLM enriches environmental understanding by constructing a semantic scene graph, inferring spatial relationships between objects to support higher-level reasoning.

Collectively, this multi-level VLM integration constitutes a significant departure from traditional navigation pipelines. Our system uses a single VLM to perform conditional detection, semantic exploration guidance, verification gating, and scene graph reasoning—functions typically handled by separate modules. This unified approach reduces architectural complexity, improves adaptability to novel goals, and enhances navigation success rates in challenging environments.

Based on the experimental design of this work, the evaluation is divided into simulation and real-world experiments. Given the inherent limitations of the simulation platform, the simulation experiments primarily focus on evaluating the performance of the Cerebrum module (\textbf{FSUNav\_Cerebrum}), with the underlying local planner implemented using the simulator’s built-in planning module. In contrast, the real-world experiments are designed to assess the overall performance of the complete cerebrum-cerebellum architecture.

\begin{figure*}
    \centering
    \includegraphics[width=1.0\linewidth]{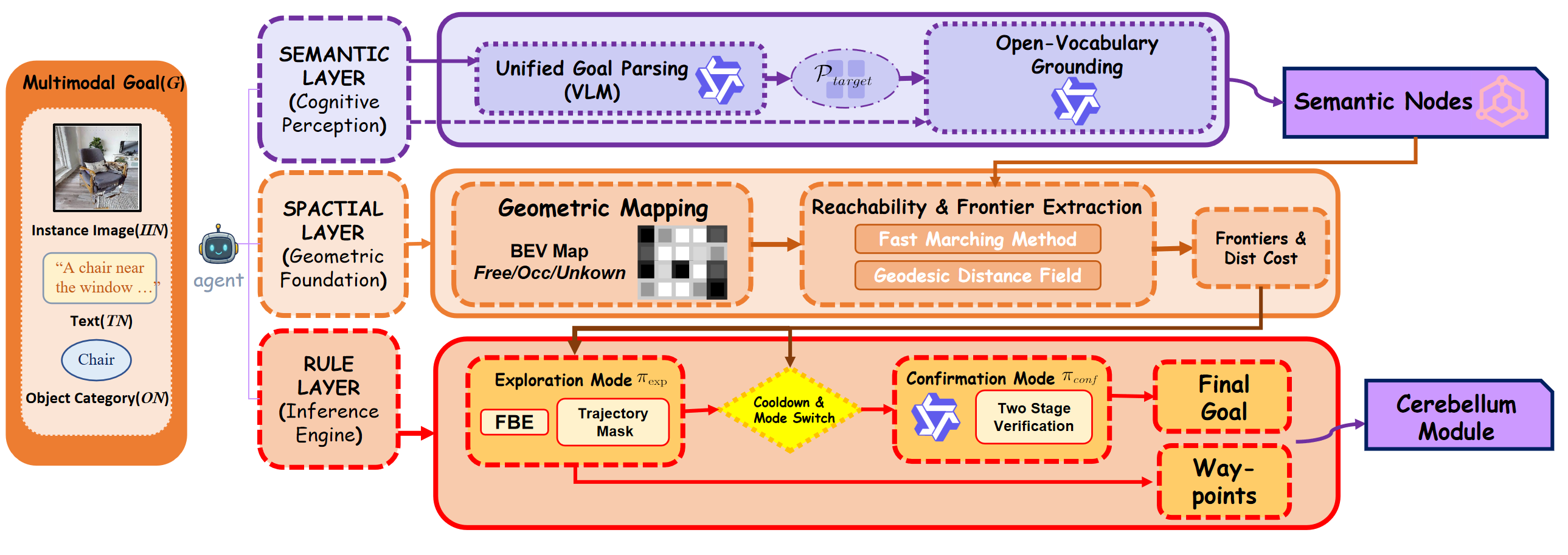}
    \caption{The overall framework of \textbf{FSUNav\_Cerebrum} is shown in the figure. A unified Vision-Language Model (VLM) serves as the core semantic engine across three Cerebrum layers. The Semantic Layer parses multimodal goals into structured target profiles and performs open-vocabulary grounding; the Spatial Layer integrates VLM-driven semantic waypoints with geometry-based frontier exploration for efficient navigation; and the Rule Layer orchestrates behavior via two-stage verification and adaptive cooldown, while also constructing a semantic scene graph. This hierarchical design enables training-free, zero-shot adaptation to heterogeneous goal-oriented navigation tasks.}
    \label{fig:placeholder}
\end{figure*}

\section{Simulation Test}

\subsection{Experimental Setup}
\label{subsec:setup}

\subsubsection{Datasets and Tasks}
We evaluate our Cerebrum module on three standard navigation tasks using the Matterport3D (MP3D~\cite{Matterport3D}) and Habitat-Matterport3D (HM3D~\cite{Ramakrishnan2021HabitatMatterport3D}) datasets: Object-Goal Navigation (ON), Instance Image-Goal Navigation (IIN) and Text-Goal Navigation (TN).

\subsubsection{Evaluation Metrics}
We employ two standard metrics for evaluation:

\begin{itemize}
    \item \textbf{Success Rate (SR)}: the percentage of episodes in which the agent successfully stops within 1 meter of the target location. This metric measures overall task completion ability.
    
    \item \textbf{Success weighted by Path Length (SPL)}: a composite metric that accounts for both success and path efficiency. It is defined as:
    \begin{equation}
        \text{SPL} = \frac{1}{N} \sum_{\mathrm{i}=1}^{N} S_\mathrm{i} \cdot \frac{l_\mathrm{i}}{\max(p_\mathrm{i}, l_\mathrm{i})},
    \end{equation}
    where $S_\mathrm{i}$ is a binary indicator of success for episode $\mathrm{i}$, $l_\mathrm{i}$ is the shortest path length from the start point to the target, and $p_\mathrm{i}$ is the actual path length taken by the agent. SPL penalizes unnecessarily long trajectories.
\end{itemize}

\subsection{Compared Methods}
We compare our method against a comprehensive set of state-of-the-art approaches across the three navigation tasks. For ObjectNav (ON) on HM3D, we consider supervised or training-based methods: ZSON~\cite{Majumdar2022ZSONZO}, PSL~\cite{Sun2024PrioritizedSL}, GOAT~\cite{Chang2023GOATGT}, and PEANUT+LOG~\cite{Qin_2025_ICCV}, as well as the strongest training-free approaches: ESC~\cite{Zhou2023ESCEW}, OpenFMNav~\cite{Kuang2024OpenFMNavTO}, VLFM~\cite{Yokoyama2023VLFMVF}, SG-Nav~\cite{yin2024sg}, CogNav~\cite{Cao2024CogNavCP}, REST~\cite{Xiao2026RESTRH}, and the universal training-free baseline UniGoal~\cite{Yin2025UniGoalTU}. For Instance Image Navigation (IIN), we compare with supervised methods IEVE~\cite{Lei2024InstanceAwareEF}, PSL~\cite{Sun2024PrioritizedSL}, and GOAT~\cite{Chang2023GOATGT}, together with UniGoal~\cite{Yin2025UniGoalTU} as the universal training-free baseline. For Target Navigation (TN), due to the lack of dedicated training‑free systems, we compare against the strongest available supervised and universal methods, including PSL~\cite{Sun2024PrioritizedSL}, GOAT~\cite{Chang2023GOATGT}, and UniGoal~\cite{Yin2025UniGoalTU}. For Open-Vocabulary Navigation (OV), we compare with supervised methods MTU3D~\cite{Zhu2025MoveTU} and VISOR~\cite{Taioli2026VISORVS}, alongside the universal training-free baseline UniGoal~\cite{Yin2025UniGoalTU}. These baselines span supervised, task‑specific, zero‑shot, and universal paradigms, providing a broad and rigorous landscape for evaluating the performance, generalization, and robustness of our proposed unified training‑free navigation framework. Our agent was implemented within the Habitat Simulator \cite{Savva2019HabitatAP}, receiving RGB-D observations at a resolution of $640 \times 480$ and a $79^\circ$ horizontal field of view (HFOV). The VLM, Qwen3-VL-32B-Instruct \cite{Qwen3-VL}, was deployed using Ollama, and all experiments were conducted on a workstation equipped with two NVIDIA GeForce RTX 4090 GPUs.

In this section, we conduct extensive experiments to validate the effectiveness of our proposed method.

\begin{table*}[t]
\centering
\caption{\textbf{Performance comparison on MP3D and HM3D datasets.}
We evaluate methods on ON, IIN, TN, and OV tasks.
Results of REST are reported from its original paper, which only provides overall ObjectNav results without task decomposition.}
\label{tab:main_results}
\small
\setlength{\tabcolsep}{6.5pt}
\begin{tabular}{@{}l*{5}{cc}*{2}{c}@{}}
\toprule
\multirow{2}{*}{Method}
& \multicolumn{2}{c}{\textbf{ON (MP3D)}}
& \multicolumn{2}{c}{\textbf{ON (HM3D)}}
& \multicolumn{2}{c}{\textbf{IIN (HM3D)}}
& \multicolumn{2}{c}{\textbf{TN (HM3D)}}
& \multicolumn{2}{c}{\textbf{OV (HM3D)}}
& \multirow{2}{*}{\textbf{Training-Free}}
& \multirow{2}{*}{\textbf{Universal}} \\
\cmidrule(lr){2-3} \cmidrule(lr){4-5} \cmidrule(lr){6-7}
\cmidrule(lr){8-9} \cmidrule(lr){10-11}
& SR & SPL & SR & SPL & SR & SPL & SR & SPL & SR & SPL & & \\
\midrule

ZSON~\cite{Majumdar2022ZSONZO} & 15.3 & 4.8 & 25.5 & 12.6 & -- & -- & -- & -- & -- & -- & No & No \\
IEVE~\cite{Lei2024InstanceAwareEF} & -- & -- & -- & -- & 70.2 & 25.2 & -- & -- & -- & -- & No & No \\
PEANUT+LOG~\cite{Qin_2025_ICCV} & 40.3 & 15.4 & 64.3 & 34.2 & -- & -- & -- & -- & -- & -- & No & No \\
MTU3D~\cite{Zhu2025MoveTU} & -- & -- & -- & -- & -- & -- & -- & -- & 40.8 & 12.1 & No & No \\

VISOR~\cite{Taioli2026VISORVS} & -- & -- & -- & -- & -- & -- & -- & -- & 28.48 & 17.26 & No & No \\
\midrule

PSL~\cite{Sun2024PrioritizedSL} & -- & -- & 42.4 & 19.2 & 23.0 & 11.4 & 16.5 & 7.5 & -- & -- & No & Yes \\
GOAT~\cite{Chang2023GOATGT} & -- & -- & 50.6 & 24.1 & 37.4 & 16.1 & 17.0 & 8.8 & -- & -- & No & Yes \\

\midrule

ESC~\cite{Zhou2023ESCEW} & 28.7 & 14.2 & 39.2 & 22.3 & -- & -- & -- & -- & -- & -- & Yes & No \\
OpenFMNav~\cite{Kuang2024OpenFMNavTO} & 37.2 & 15.7 & 54.9 & 24.4 & -- & -- & -- & -- & -- & -- & Yes & No \\
VLFM~\cite{Yokoyama2023VLFMVF} & 36.4 & 17.5 & 52.5 & 30.4 & -- & -- & -- & -- & -- & -- & Yes & No \\
SG-Nav~\cite{yin2024sg} & 40.2 & 16.0 & 54.0 & 24.9 & -- & -- & -- & -- & -- & -- & Yes & No \\
CogNav~\cite{Cao2024CogNavCP} & 46.6 & 16.1 & 72.5 & 26.2 & -- & -- & -- & -- & -- & -- & Yes & No \\
REST~\cite{Xiao2026RESTRH} & -- & -- & 57.3 & 33.4 & -- & -- & -- & -- & -- & -- & Yes & No \\

\midrule

UniGoal~\cite{Yin2025UniGoalTU} & 41.0 & 16.4 & 54.5 & 25.1 & 60.2 & 23.7 & 20.2 & 11.4 & -- & -- & Yes & Yes \\

\textbf{FSUNav\_Cerebrum} & \textbf{48.75} & \textbf{26.93} & \textbf{76.2} & \textbf{40.49} & \textbf{72.6} & \textbf{34.88} & \textbf{39.8} & \textbf{21.58} & \textbf{56.87} & \textbf{32.59} & Yes & Yes \\

\bottomrule
\end{tabular}
\end{table*}

\subsection{Comparison with State-of-the-art}

As shown in Table~\ref{tab:main_results}, we compare our \textbf{FSUNav\_Cerebrum} against a wide range of state‑of‑the‑art methods, including supervised approaches (ZSON~\cite{Majumdar2022ZSONZO}, IEVE~\cite{Lei2024InstanceAwareEF}, PEANUT+LOG~\cite{Qin_2025_ICCV}), universal supervised methods (PSL~\cite{Sun2024PrioritizedSL}, GOAT~\cite{Chang2023GOATGT}), training‑free methods (ESC~\cite{Zhou2023ESCEW}, OpenFMNav~\cite{Kuang2024OpenFMNavTO}, VLFM~\cite{Yokoyama2023VLFMVF}, SG-Nav~\cite{yin2024sg}, CogNav~\cite{Cao2024CogNavCP}, REST~\cite{Xiao2026RESTRH}), and the recent training‑free universal method UniGoal~\cite{Yin2025UniGoalTU}. Our method consistently achieves superior performance across all tasks and datasets. On the MP3D Object Navigation (ON) benchmark, \textbf{FSUNav\_Cerebrum} attains a Success Rate (SR) of 48.75\% and an SPL of 26.93\%, outperforming all prior works—including the strongest training‑free competitor CogNav (46.6\% SR, 16.1 SPL) and the supervised PEANUT+LOG (40.3\% SR, 15.4 SPL). On the more challenging HM3D ON task, our method achieves 76.2\% SR and 40.49\% SPL, substantially surpassing both supervised methods (e.g., PEANUT+LOG: 64.3\% SR, 34.2 SPL) and training‑free methods (e.g., CogNav: 72.5\% SR, 26.2 SPL). For the Instance Image Navigation (IIN) task on HM3D, \textbf{FSUNav\_Cerebrum} reaches 72.6\% SR and 34.88\% SPL, significantly outperforming the supervised IEVE (70.2\% SR, 25.2 SPL) and the universal UniGoal (60.2\% SR, 23.7 SPL). On the hardest Target Navigation (TN) benchmark, our method achieves 39.8\% SR and 21.58\% SPL, nearly doubling the performance of prior universal methods such as UniGoal (20.2\% SR, 11.4 SPL) and GOAT (17.0\% SR, 8.8 SPL). Notably, \textbf{FSUNav\_Cerebrum} is both training‑free and universal, demonstrating strong generalization without any task‑specific fine‑tuning. Furthermore, we evaluate open‑vocabulary (OV) navigation on HM3D, where our method achieves 56.87\% SR and 32.59\% SPL, surpassing existing OV methods such as MTU3D~\cite{Zhu2025MoveTU} (40.8\% SR, 12.1 SPL) and VISOR~\cite{Taioli2026VISORVS} (28.48\% SR, 17.26 SPL). The last two columns confirm that \textbf{FSUNav\_Cerebrum} is the only method that is both training‑free and universal while explicitly supporting open‑vocabulary navigation, highlighting its ability to handle arbitrary object categories beyond a fixed training set. These results collectively demonstrate the effectiveness and versatility of our approach across diverse navigation scenarios.

\begin{figure}
    \centering
    \includegraphics[width=0.8\linewidth]{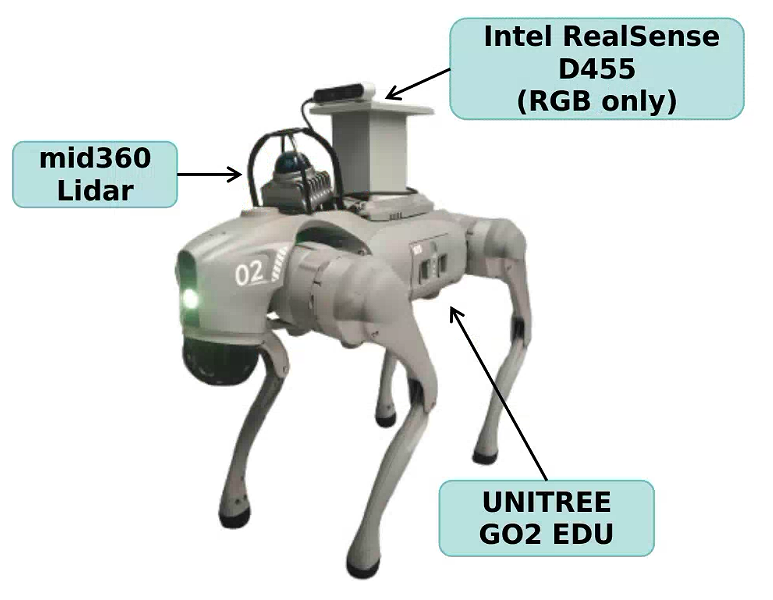}
    \caption{In our real-world experimental setup, we deployed the framework on a Unitree Go2 EDU quadruped robot. A custom 3D-printed mount was used to attach an Intel RealSense D455 camera to the robot for capturing real-time RGB observations.}
    \label{fig:placeholder}
\end{figure}

\section{Real-world Test}

As shown in Figure~\ref{fig:placeholder}, this paper conducts physical experiments using the Unitree Go2 EDU quadruped robotic platform. \textbf{Experiments involving the humanoid robot Unitree G1 and the four-wheel mobile robot will be reported in subsequent sections.} The robotic platform is equipped with an Intel RealSense D455 depth camera, which is rigidly fixed to the front of the robot via a custom-designed 3D-printed mount, capturing RGB images at a resolution of $640 \times 480$. It is worth noting that this work utilizes only RGB information as input, from which calibrated depth maps are estimated using the VGGT model. This design aims to verify the feasibility of achieving closed-loop navigation using only RGB perception, demonstrating that the proposed method does not rely on dedicated depth sensors, thereby enhancing its generalizability across different platforms and deployment scenarios.

In contrast to simulation environments, to ensure the stability and reliability of perception under high-speed motion during physical experiments, we incorporate a Livox MID-360 LiDAR as a complementary sensing unit. This LiDAR is employed to construct a BEV environmental map and simultaneously supports real-time collision avoidance in the underlying cerebellum module. This configuration enhances robustness against dynamic scenarios through heterogeneous sensor fusion while providing a stable local environmental representation for the high-level navigation module.

The real-time acquisition of the robot's pose and the execution of control commands from the cerebellum module are realized via the official Unitree Python SDK. Communication between the robot and the remote workstation is established over a wireless local area network (WLAN), enabling low-latency transmission of both control signals and sensory data.

\begin{figure*}
    \centering
    \includegraphics[width=1.0\linewidth]{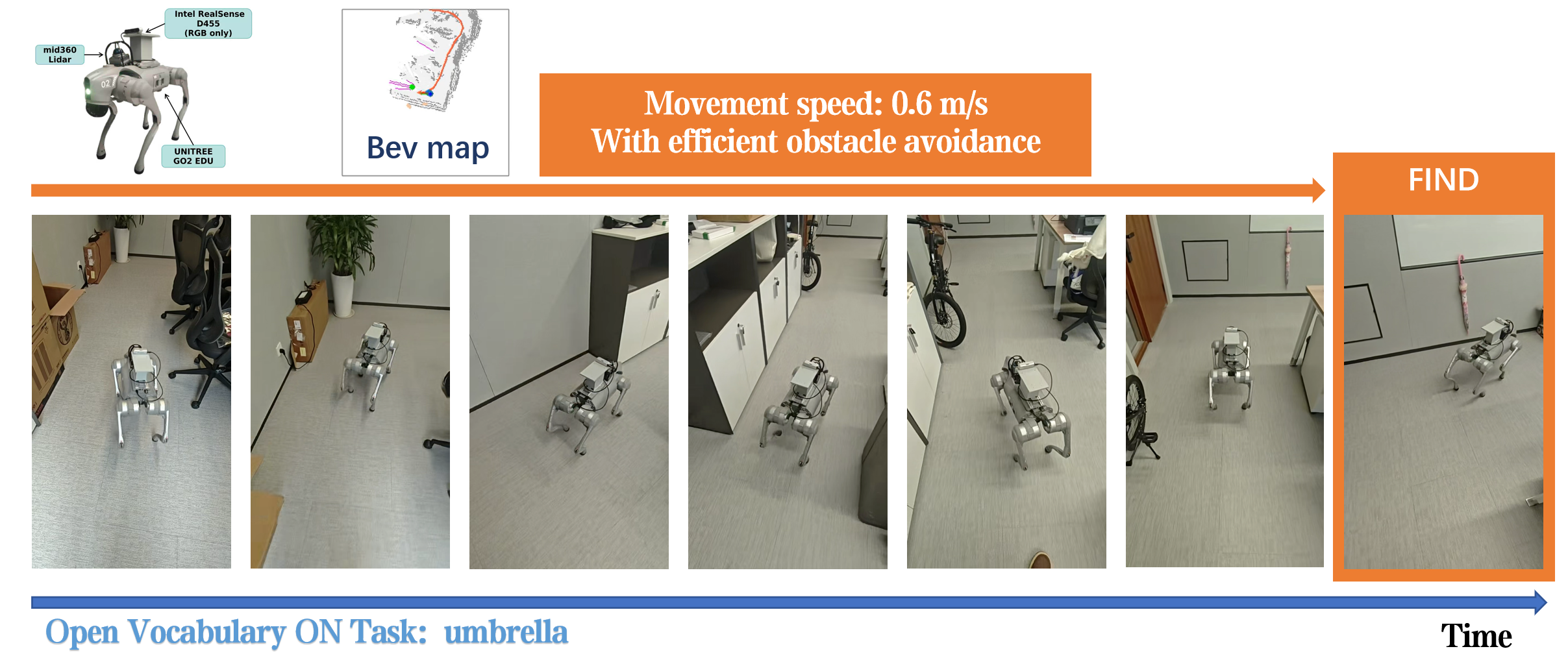}
    \caption{Under a maximum locomotion speed of 0.6 m/s, the quadruped robot successfully completed the open‑vocabulary object goal navigation task targeting “umbrella.” During the experiment, the robot not only demonstrated efficient mobility but also exhibited real‑time dynamic obstacle avoidance capabilities, achieving safe approach and accurate identification of the target object in unstructured environments. Additional task types for the quadruped robot, along with physical experiments involving the humanoid robot G1 and wheeled mobile robots, are planned to be updated in subsequent versions.}
     \label{fig:result}
\end{figure*}

As shown in Figure \ref{fig:result}, under a maximum locomotion speed of 0.6 m/s, the quadruped robot successfully completed the open‑vocabulary object goal navigation task targeting “umbrella.” Experimental results demonstrate that during high-speed motion, the robot not only exhibits efficient mobility but also possesses real‑time dynamic obstacle avoidance capabilities, enabling safe approach and accurate identification of the target object in unstructured environments. The above experiments validate the effectiveness of the proposed framework in balancing locomotion speed and task success rate. Additional task types for the quadruped robot, along with physical experimental results involving the humanoid robot G1 and wheeled mobile robots, are planned to be updated in subsequent versions.

\section{Conclusion}
Based on the FSUNav framework proposed in this paper, we systematically address key bottlenecks in existing vision-language navigation methods through the efficient collaboration of the Cerebrum and Cerebellum modules, including heterogeneous robot compatibility, real-time performance, navigation safety, and open-vocabulary semantic generalization. The Cerebellum module, built on deep reinforcement learning, achieves a universal local planner that supports cross-platform deployment across various robotic platforms such as wheeled, quadruped, and humanoid robots, significantly improving navigation efficiency and collision avoidance safety. The Cerebrum module leverages a vision-language model to construct a unified three-layer reasoning architecture, enabling zero-shot understanding and localization of object categories, instance images, and natural language instructions, achieving leading performance on multiple standard datasets. Real-world experiments further validate the robustness and practicality of the framework in complex dynamic environments. Overall, FSUNav provides a systematic solution that is efficient, safe, and highly generalizable for the practical deployment of vision-language navigation on heterogeneous robotic platforms.


%

\ifCLASSOPTIONcaptionsoff
  \newpage
\fi



%
\bibliographystyle{IEEEtran}
\bibliography{mybib}

\end{document}